# MACHINE LEARNING MODEL TO PREDICT SOLAR RADIATION, BASED ON THE INTEGRATION OF METEOROLOGICAL DATA AND DATA OBTAINED FROM SATELLITE IMAGES

Luis Eduardo Ordoñez Palacios[1], Víctor Bucheli Guerrero[2], Hugo Ordoñez[3]

luis.ordonez.palacios@correounivalle.edu.co (author for correspondence),
victor.bucheli@correounivalle.edu.co, hugoordonez@unicauca.edu.co

[1][2] Escuela de Ingeniería de Sistemas y Computación (EISC),
Universidad del Valle, Santiago de Cali, Colombia

[3] Facultad de Electrónica y Telecomunicaciones, Departamento de Sistemas,
Universidad del Cauca, Popayán, Colombia

**Abstract:** Knowing the behavior of solar radiation at a geographic location is essential for the use of energy from the sun using photovoltaic systems; however, the number of stations for measuring meteorological parameters and for determining the size of solar fields in remote areas is limited. In this work, images obtained from the GOES-13 satellite were used, from which variables were extracted that could be integrated into datasets from meteorological stations. From this, 3 different models were built, on which the performance of 5 machine learning algorithms in predicting solar radiation was evaluated. The neural networks had the highest performance in the model that integrated the meteorological variables and the variables obtained from the images, according to an analysis carried out using four evaluation metrics; although if the rRMSE is considered, all results obtained were higher than 20%, which classified the performance of the algorithms as fair. In the 2012 dataset, the estimation results according to the metrics MBE, $R^2$, RMSE, and rRMSE corresponded to -0.051, 0.880, 90.99 and 26.7%, respectively. In the 2017 dataset, the results of MBE, $R^2$, RMSE, and rRMSE were -0.146, 0.917, 40.97 and 22.3%, respectively.

**Keywords:** Satellite images, GOES-13, Meteorological stations, Solar Radiation, Sunshine, Predictive model.

# 1. Introduction

The electromagnetic radiation coming from the Sun is an inexhaustible source of energy that can be exploited to produce electricity [1]; sunlight can be transformed into electric power using photovoltaic systems and the heat of the Sun; however, the Sun radiation that arrives at the surface of Earth undergoes a weakening process due to several dispersion, reflection and absorption factors, as can be observed in [2]. Therefore, it is important to know the behavior of Sun radiation in a specific geographic location to determine the size of systems that allow its utilization.

There are various types of instruments that measure solar radiation [3] and some organizations interested in monitoring the meteorological conditions of a region have at their disposal a limited number of measurement stations to make observations regarding the behavior of the characteristics that determine the weather. In Colombia, the Institute of Hydrology, Meteorology and Environmental Studies (IDEAM), is a government entity dependent on the Minister of Environment and Sustainable Development, and is in charge of handling the scientific information related to the environment; likewise, there are other entities [4] [5] in the public and private spheres that also make these kinds of measurements.

Although there are various organizations that monitor weather conditions, the number of ground-based measurement stations [6], is not sufficient to cover the entire Colombian geography, since they imply additional costs for maintenance and surveillance [7]. Additionally, many stations do not capture radiation values due to the adverse conditions they are exposed to [8]. Therefore, several researchers have developed different physical, statistical, and artificial intelligence models to estimate solar radiation supported by terrestrial and satellite measurement instruments.

According to the research of Suarez Vargas [9], physical and statistical models are based on an energy balance between the radiation reaching the top of the atmosphere and the radiation reflected by the satellite. On the other hand, physical models use parameters of absorption, spreading, cloud albedo and superficial albedo, although the difficulty of these models lies in knowing these atmospheric values at a local level, according to the research of Zarzalejo et al. [10]; in the case of statistical models, regressions are used between the radiometric measurements on the surface and the information recorded by the satellite, as observed in the work of Poveda Matallana [11].

Many studies have used satellite images to calculate solar radiation; the work of Docel Ballén [12] used the Heliostat 1 algorithm to estimate solar irradiance using images from the GOES satellite over the Cundiboyacense region in Colombia; in a similar way, the research of Albarelo et al. [13] used a modification of the Heliosat-II method, developed to process images from the Meteostat satellite, for its use with images from the GOES satellite in solar radiation estimation over the French Guyana; in the same way, the work of Pagola [14] implemented a combination between the Heliostat 1 and Heliostat 2 methods to estimate solar radiation in Spain, using images coming from the second generation Meteostat satellite. The Heliostat method has also been used to estimate solar radiation in research developed by Hammer et al. [15] [16], Kallio-Myers et al. [17], Lorenz et al. [18] [19], Rigollier et al. [20] and Zarzalejo et al. [10].

Other researchers have used the Angström-Prescott model for the calculation of solar radiation on the surface using radiometric stations; this model relates the monthly average solar radiation with the solar radiation on a clear day using the hours of daily sunlight, according to the research of Prescott [21]. Research by Poveda Matallana [11] validated the solar radiation on the surface over Orinoquía from images obtained from the GOES satellite. The research indicated that the Angström-Prescott coefficients depend on geographic and climatic parameters and on dynamic, spatial and physical properties of the atmosphere, which explains the need for radiation data and sunshine obtained by measurement stations on earth. In the same way, the work of Guzmán M. et al. [22] utilized the Angström-Prescott coefficients to estimate global solar radiation using sunshine data in the coffee zone in Colombia.

In this review, we also found studies that used artificial intelligence techniques to estimate solar radiation, based on historical meteorological data and in some studies, images provided by geostationary satellites; the studies of Eissa et al. [23], Hammer et al. [15], Linares-Rodriguez et al. [24] [25], Martín Pomares et al. [26], Mazorra Aguiar et al. [27], Ordoñez-Palacios et al. [28], Gürel et al. [29], Jumin et al. [30] and Ağbulut et al. [31], used machine learning algorithms; on the other hand, the studies of Alzahran et al. [32], Jiang et al. [33] [34], Kaba et al. [35], and Chandola et al. [36] utilized deep learning techniques.

In this work, images obtained from the GOES-13 meteorological satellite from 2012 and 2017 were processed [37] using Python geographic information libraries (Rasterio, Pyproj). It is important to note that (i) a model was built based on the Ångström-Prescott method for the estimation of solar radiation from data observed on the ground and data extracted from satellite images; (ii) a model was proposed to predict solar radiation using variables extracted from the images, such as reflectance, cloudiness index, number of hours of daily sunlight and solar radiation at the edge of the atmosphere; (iii) a model was implemented to estimate solar radiation incorporating two dimensions of information: variables obtained from the images and meteorological variables (direction and speed of the wind, temperature, rain and humidity) obtained from measurement stations on land, and the integration was implemented through repetitive routines in Python, validating the date and time of each observation; and (iv) it is possible to predict solar radiation at any geographic location on the planet, if you utilize the satellite images for that specific place.

This paper includes the following sections: methodology, the results, discussion and future work, and finally, conclusions.

## 2. Materials and methods

This section formulates the questions of interest that guided this investigation, the information sources that provided the data, the way it was processed, the methods that were applied to estimate solar radiation and the architecture of the regression model.

### 2.1 Questions of interest

Solar energy is a renewable resource that can be transformed into electric power using photovoltaic systems; therefore, it is fundamental to understand its behavior in regard to the radiation levels reaching the Earth, either by measurement instruments, mathematical calculations from satellite images or predictions using artificial intelligence models.

Considering that the number of measurement stations in Colombia is insufficient and that they are exposed to constant risks, such as to crime issues, geographic location and adverse climatic conditions, it is necessary to evaluate existing methods for estimation; therefore, questions such as the following are important: what is the process to calculate solar radiation from images and what other variables are obtained? What techniques of machine learning can be utilized to predict solar radiation and which have a better performance? What are the results obtained by utilizing meteorological variables extracted from the images or the integration of both? These questions will be answered throughout this paper.

**2.2 Information sources**

For this research, historical images of the visible channel were utilized, obtained from the GOES-13 meteorological satellite during 2012 and 2017; additionally, two sets of data from air quality stations (Republic of Argentina School –ERA– and Compartir) of the Administrative Department of Environment Management from the Mayoralty of Cali and a set of data of daily sunshine from the Univalle station provided by the Institute of Hydrology, Meteorology and Environmental Studies (IDEAM) were obtained.

The images were captured by the satellite every hour and a half, and for this research, the pictures obtained from 6 am to 6 pm were utilized; however, it is important to highlight that only a few days throughout the year account for all the images; on the other hand, both sets of air quality data (ERA and Compartir) include historical observations of wind speed and direction, temperature, rain, humidity and solar radiation recorded every hour. Likewise, it is worth mentioning that, in the set of sunshine data from the IDEAM, there were some days of the year that did not have a value for this variable. Tables 1, 2 and 3 show the metadata of the images and sets of data utilized.

**Table 1**. Satellite images.

| ID | Year | Amount of images | Size | Total |
|---|---|---|---|---|
| 1 | 2012 | 1991 out of 4758 | 9.6 MB | 19.11 GB |
| 2 | 2017 | 2135 out of 4745 | 9.6 MB | 20.50 GB |

**Table 2**. Datasets from the air quality stations.

| ID | Station | Latitude | Longitude | Year | Registers |
|---|---|---|---|---|---|
| 1 | ERA | 3.44779 | -76.51918 | 2012 | 8634 |
| 2 | Compartir | 3.42823125782003 | -76.46654484665319 | 2017 | 8166 |

**Table 3**. Datasets of daily sunshine.

| ID | Station | Latitude | Longitude | Year | Registers | Percentaje |
|---|---|---|---|---|---|---|
| 1 | Univalle | 3.3780 | -76.53388889 | 2012 | 329 out of 366 | 89.9% |
| 2 | Univalle | 3.3780 | -76.53388889 | 2017 | Not available | 0% |

The procedure to obtain the pictures starts with the register of a user and login on the website of the electronic library CLASS of the National Office of Oceanic and Atmospheric Administration (NOAA) [38]; the product GOES Satellite Data – Imager (GVAR_IMG) is selected, and the area, period of time, coverage and satellite are determined. On the next website, the dataset is chosen, which can be approximately 60 packages for a specific year. Considering that it is possible to request only one at a time, the download options must be chosen and must indicate that the images will be used for educational purposes. The request of all the packages can take approximately 6 hours depending on the connection speed.

In this research, a web tracker was developed for the automatic download of satellite images, files with a goal extension were filtered, and images from channels other than the visible channel (BAND_02, BAND_03, BAND_04, BAND_05 and BAND_06) and the images corresponding to the first quarter of the hour, considering that the satellite provides images every half an hour, were used. The web tracker downloaded 25 to 52 valid images automatically for this research, in an approximate time of 4 hours per link, with a residential connection of 75 MB, though the connection speed depended on server availability.

The images obtained from the satellite were processed with the software Weather and Climate Tools from the NOAA (WTC) [39] to visualize and export the data to the NetCDF format, which was used for the model developed in the programming language Python for the calculation of solar irradiance using the Angström-Prescott method. This process can take approximately 4 hours, depending on the characteristics of the computer equipment and, in some cases, on the export of images without data by the software.

### 2.3 Angström-Prescott method for the calculation of irradiance

Estimation of solar radiation ($H$) of a geographic location using the Angström-Prescott model requires calculation of the coefficients a and b, the cloudiness index ($n_c$) and solar radiation at the edge of the atmosphere ($H_{ext}$), according to research by Poveda Matallana [11]. The coefficients are obtained from the relationship between the number of hours of sunlight and global solar radiation, with both parameters captured on the surface, according to Guzmán M. et al. [22]. The equation for the calculation of solar radiation is:

$$H = [a + b*(1 - n_c)] * H_{ext} \tag{1}$$

In this sense, to obtain the cloudiness index, first, it is necessary to transform the images to the TIFF format using the libraries Rasterio and PyProj from Python. Later, the digital level of the image is obtained, thus, along with the satellite calibration coefficient [40], the nominal reflectance is calculated ($R_{prev}$) (2); likewise, the monthly correction factor ($C$) [41] of the satellite and the nominal reflectance allow us to calculate the rear reflectance ($R_{post}$) (3), which makes it possible to obtain the pixel reflectance ($R_p$) (4) of every image, considering the calculation of the astronomical variables, distance from the Earth to the Sun ($r$) and zenith angle ($\theta_z$). The model also requires the calculation of other variables, such as solar declination, the equation of time, true solar time and astronomical length of the day.

$$R_{prev} = k\ (nd - 29) \qquad (2)$$

$$R_{post} = C * R_{prev} \qquad (3)$$

$$R_p = (R_{post} * r^2) \div cos(\theta_z) \qquad (4)$$

The reflectance calculation process can take approximately 4 hours on a computer with a 4 physical core processor (8 logical) and 12 GB RAM memory. With the data obtained, the cloudiness index can be calculated ($n_c$) (5), finding the maximum ($R_{max}$) and minimum ($R_{min}$) reflectance for each hour of the day, although according to Laguarda [42], the original value must be taken from the minimum reflectance and the 80% of the maximum reflectance; however, the values of the cloudiness index must be between 0 and 1; therefore, they must be adjusted if they are out of the domain.

$$n_c = (R_p - R_{min}) \div (R_{max} - R_{min}) \qquad (5)$$

Estimating solar radiation from satellite imagery requires solar radiation and brightness data captured by ground-based measurement stations; however, the air quality datasets from the DAGMA (ERA and Compartir) contain only solar radiation. Consequently, sunshine data were requested from a nearby IDEAM station. In this case, they provided data from the Univalle station, located 9.3 km from the Compartir station, and 7.8 km from the ERA station. Later, the sunshine data were integrated into the ERA station, although it was not possible to do the same for the data from the Compartir station, since IDEAM does not have data on sunshine for 2017. With radiation and sunshine data integrated into the ERA station dataset, the monthly coefficients $a$ and $b$ were calculated for 2012, which allowed the calculation of solar radiation with Equation (1).

## 2.4 Machine Learning Algorithms

The regression models for the prediction of solar radiation were developed in the python programming language on the Jupyter Notebook web-based interactive platform; the MinMaxScaler method was used to normalize the data between 1 and 0. In addition, machine learning techniques were also included: multiple linear regression, XGBoost for regression, artificial neural networks and the assembly methods: gradient boosting and random forest. In the matter of neural networks, the Multilayer Perceptron was used and tested with the 2012 data, about 4000 different configurations in 7 hyperparameters, establishing a search space of 3 to 7 different values in each of the following hyperparameters: solver, hidden_layer_sizes, alpha, activation, learning_rate, max_iter, n_iter_no_change; as for the alpha parameter, a range of values between: 1e-5 and 1e-3 was set using the loguniform function of the scipy.stats library.

Subsequently, the hyperparameter tuning methods of the machine learning library scikit_learn were experimented with: grid search (GridSearchCV) and random search: (RandomizedSearchCV); both methods use cross-validation to improve the estimated performance of a model, using the RepeatedKFold technique; in this work, the data was randomly divided into 10 subsets in each iteration. In the same way, the loss function that was tried to be optimized is the mean square error, understanding that the reduction of this metric improves the model estimates.

Hyperparameter adjustment with random search was used because it obtained results as precise as those obtained with grid search, but with a considerable reduction in time, thanks to the sampling of the hyperparameters in the defined distribution. The search space of the hyperparameters and the values of the best estimator are presented in Table 4.

**Table 4**. Search space and values of the best model.

| ID | Hyperparameter | Search space | Found value |
|---|---|---|---|
| 1 | Optimization algorithm | ["lbfgs", "adam", "sgd"] | Adam |
| 2 | Number of hidden layers | [[125, 100, 75, 50, 10], [100, 75, 50, 10], [100, 75, 50], [100, 75]] | [125, 100, 75, 50, 10] |
| 3 | L2 regularization term | loguniform(1e-5, 1e-3) | 0.0001462317189 |
| 4 | Activation function | ["identity", "logistic", "tanh", "relu"] | Relu |
| 5 | Learning rate | ["constant", "adaptive", "invscaling"] | Invscaling |
| 6 | Maximum number of iterations | [200, 1000, 5000, 10000] | 1000 |
| 7 | Maximum number of epochs | [5, 10, 15, 20, 25, 30, 40] | 30 |

Three regression models were built to estimate solar radiation: the first (M1) used meteorological variables (wind speed and direction, temperature, rain, and humidity), the second (M2) only included the variables obtained from the satellite images (reflectance, cloudiness index, solar radiation at the edge of the atmosphere and the theoretical number of hours of solar brightness) and the third (M3) integrated both groups of variables. Table 5 describes each of the variables.

**Table 5**. Description of variables.

| ID | Source | Variable | Description |
|---|---|---|---|
| 1 | Measurement station | Wind speed | It refers to the displacement of air at a point and at a given moment; it is measured in meters per second (m/s). |
| 2 | | Wind direction | Indicates the direction in degrees (0-360) from where the wind is coming. |
| 3 | | Temperature | It is related to the notion of heat in the atmosphere and is measured in Celsius degrees. |
| 4 | | Rain | It is defined as the amount of water that falls per unit of time in a given place, measured in millimeters (mm). |
| 5 | | Humidity | It refers to the vapor present in the atmosphere. |
| 6 | | Solar Radiation (Variable objective) | Energy flow received from the Sun in the form of electromagnetic waves, measured in $W/m^2$. |
| 7 | Satellite images | Reflectance | It corresponds to the value of the solar radiation that is reflected by the clouds. |
| 8 | | Cloudiness index | It is a value related to the cloudiness conditions of clear, partly cloudy, and cloudy skies (0-1). |
| 9 | | Solar radiation at the edge of the atmosphere | It is the value of the electromagnetic radiation emitted by the Sun, before entering the atmosphere, measured in $W/m^2$. |
| 10 | | Number of hours of solar brightness | It is the time in hours during which the sun has an effective solar brightness. |

**2.5 Evaluation metrics**

In this paper, the performance of the machine learning algorithms used in the research is evaluated, and for this purpose, the following metrics were used: mean bias error (MBE), determination coefficient ($R^2$), root mean square error (RMSE) and relative root mean square error (rRMSE). Their equations, descriptions, and performance criteria for each evaluation metric are given in Table 6.

Table 6. Evaluation metrics.

| Metrics | Equation | Description | Performance Criteria |
|---|---|---|---|
| MBE | $\frac{1}{n}\sum_{i=1}^{n}(y_i - x_i)$ | MBE provides insight into the long-term performance of models. [43] | The closer the MBE value to zero, the better the estimation result. [44] |
| $R^2$ | $1 - \frac{\Sigma(y_i - x_i)^2}{\Sigma(x_i - \bar{x}_i)^2}$ | $R^2$ is used to determine how well the regression line approximates the actual data points. [45] | $R^2$ value changes between 0 and 1, and the closer this value is to 1, the better the performance of the model. [45] |
| RMSE | $\sqrt{\frac{1}{n}\sum_{i=1}^{n}(y_i - x_i)^2}$ | RMSE represents the difference between the estimated and observed values. [46] | RMSE takes positive values and the closer the RMSE value to zero, the better the estimation result. [47] |
| rRMSE | $\frac{\sqrt{\frac{1}{n}\Sigma_{i=1}^{n}(y_i - x_i)^2}}{\underline{x_i}} \times 100\%$ | rRMSE provides a percentage value, the result of dividing the RMSE by the mean of the real values. [47] | An rRMSE value close to zero explains that the models perform better. According to the investigations of Ağbulut et al. [31], Fan et al. [46] and Bakay and Ağbulut [47], the success of the estimation algorithms was classified as follows: **Excellent**: rRMSE < 10%; **Good**: 10% < rRMSE <= 20%; **Fair**: 20% < rRMSE <= 30%; **Poor**: rRMSE > 30% |

In Table 6, $y_i$ are the data estimated by the models, $x_i$ are the data captured by the measurement station; $\bar{x}_i$ is the mean of the measured data, and n, represents the number of observations.

### 2.6 Model architecture

Figure 1 explains the data flow, from terrestrial and satellite information sources to solar radiation predictions by regression techniques, passing through each of the model stages.

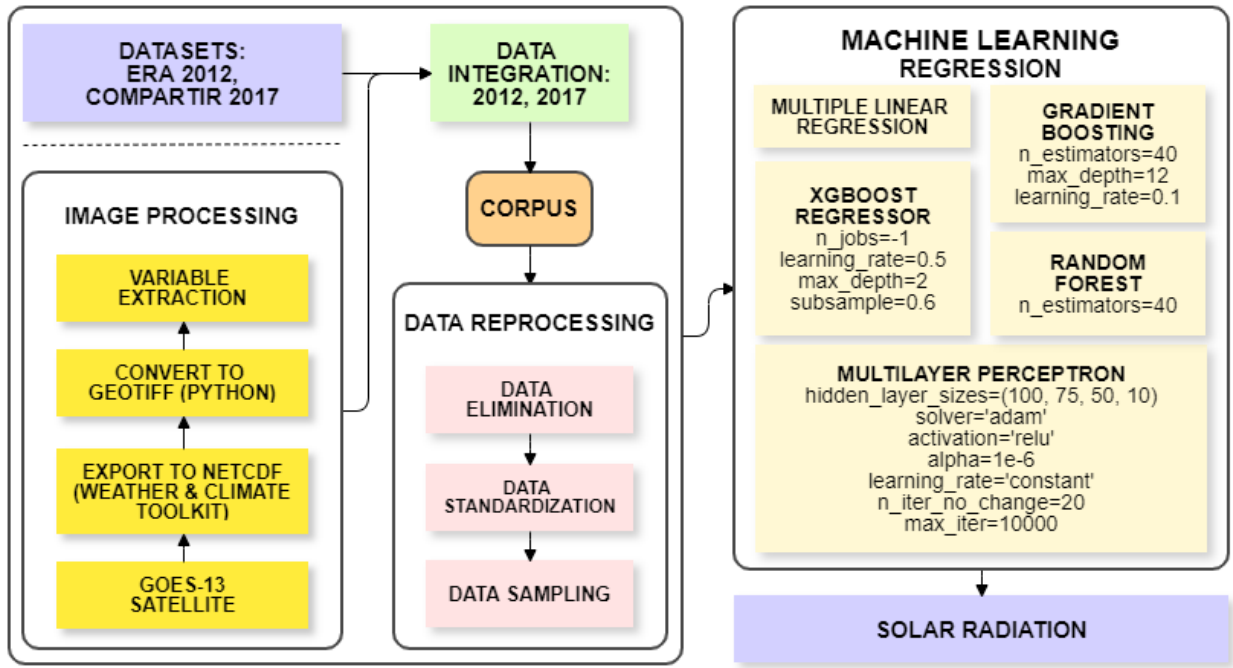

**Figure 1**. General architecture of the model.

### 3. Results

In the statistical model, regression was used to obtain the monthly Angström-Prescott coefficients for 2012, considering the relationship between radiation and solar brightness captured on the ground, which allowed the estimation of solar radiation from satellite images. For 2017, it was not possible to calculate solar radiation due to the lack of solar brightness data. Table 7 presents the values of the coefficients $a$ and $b$ and the determination coefficient $R^2$ for each month.

**Table 7**. Monthly coefficients, year 2012.

| Month | a | b | $R^2$ |
|---|---|---|---|
| 1 | 0.207 | 0.419 | 0.758 |
| 2 | 0.222 | 0.393 | 0.648 |
| 3 | 0.236 | 0.410 | 0.740 |
| 4 | 0.245 | 0.375 | 0.603 |
| 5 | 0.274 | 0.319 | 0.611 |
| 6 | 0.314 | 0.297 | 0.540 |
| 7 | 0.282 | 0.366 | 0.647 |
| 8 | 0.300 | 0.335 | 0.712 |
| 9 | 0.323 | 0.335 | 0.795 |
| 10 | 0.226 | 0.416 | 0.550 |
| 11 | 0.263 | 0.339 | 0.521 |
| 12 | 0.262 | 0.311 | 0.623 |

Figure 2 shows the scatter diagram obtained from daily solar radiation captured by the measurement station and estimated by the statistical model with a determination coefficient $R^2$ of 0.552 and a mean square error RMSE of 1062.71. The experiment was also conducted with hourly solar radiation data; however, the results showed a much lower performance.

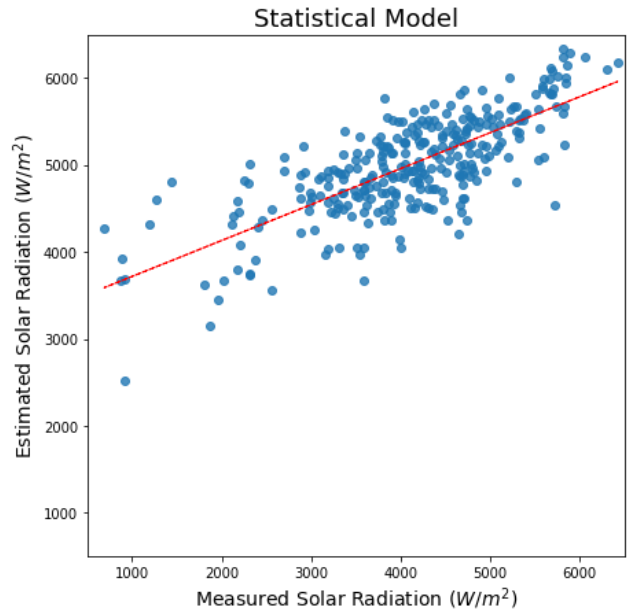

**Figure 2.** Daily solar radiation observed by the station and calculated from satellite images.

The dataset from the ERA station for 2012 was integrated into the dataset generated by processing the images of the GOES-13 satellite taken in 2012; in the same way, the dataset of the Compartir station for 2017 was integrated into the dataset of the images of the GOES-13 satellite taken in 2017. In both datasets, records were eliminated due to the nonexistence of meteorological data of the dates and times in which data from the images did exist. Table 8 presents the characteristics of the datasets used in the regression models.

**Table 8**. Datasets used in the regression.

| ID | Datasets | Registers | Eliminated | Total |
|---|---|---|---|---|
| 1 | 2012 | 1991 | 28 | 1963 |
| 2 | 2017 | 2135 | 110 | 2025 |

Tables 9 and 10 show the performance of the regression algorithms applied to the 2012 and 2017 datasets. In both cases, neural networks had the best performance; likewise, the model that integrated the meteorological variables and the variables obtained from the images had a higher performance than the other two models. It is important to point out that the machine learning models performed better than the statistical model.

The ensemble methods (gradient boosting, XGBoost regressor and random forest) with the default hyperparameters, showed results very close to 1 in training, according to the determination coefficient $R^2$, which indicated an overfitting of the algorithms. Therefore, cross-validation was used as a preventive measure against overtraining, dividing the data into 5 subsets and training the model iteratively. The results presented are the average of the values obtained in each data subset; the regularization technique was also used to artificially force the algorithm to be simpler.

The adjustment of the hyperparameters of each model was performed by random search of the scikit-learn library. A search space of between 6 and 10 different values was defined in the hyperparameters: n_estimators, max_depth, learning_rate, min_samples_leaf. In the case of the learning_rate parameter, a range of values between: 1e-3 and 1e-1 was established using the loguniform function of the python scipy library.

**Table 9**. Results obtained from the 2012 dataset.

| ID | MODEL | METRICS | MULTIPLE LINEAR REGRESSION | GRADIENT BOOSTING | XGBOOST REGRESSOR | RANDOM FOREST | NEURAL NETWORKS |
|---|---|---|---|---|---|---|---|
| M1 | METEOROLOGICAL VARIABLES | MBE | 9.467 | -0.269 | **0.171** | 0.215 | -0.808 |
| | | $R^2$ ENT | 0.564 | **0.868** | 0.863 | 0.857 | 0.858 |
| | | $R^2$ PRU | 0.559 | 0.824 | 0.817 | 0.820 | **0.849** |
| | | RMSE | 169.49 | 116.10 | 106.91 | 103.44 | **98.68** |
| | | rRMSE | 51.14 | 33.12 | 31.43 | 30.97 | **29.78** |
| M2 | VARIABLES OBTAINED FROM THE IMAGES | MBE | -0.422 | **-0.249** | -1.541 | -0.886 | 1.813 |
| | | $R^2$ ENT | 0.374 | 0.852 | **0.853** | 0.842 | 0.836 |
| | | $R^2$ PRU | 0.432 | 0.810 | 0.800 | 0.807 | **0.836** |
| | | RMSE | 192.26 | 110.97 | 106.74 | 108.47 | **103.27** |
| | | rRMSE | 58.01 | 32.56 | 31.79 | 31.58 | **31.16** |
| M3 | METEOROLOGICAL VARIABLES + | MBE | -0.248 | 0.993 | -2.968 | -2.909 | **-0.051** |
| | | $R^2$ ENT | 0.669 | 0.904 | 0.899 | 0.884 | **0.909** |

| | VARIABLES OBTAINED FROM THE IMAGES | R² PRU | 0.685 | 0.860 | 0.848 | 0.846 | **0.880** |
|---|---|---|---|---|---|---|---|
| | | RMSE | 143.19 | 96.79 | 94.87 | 98.10 | **90.99** |
| | | rRMSE | 43.21 | 28.36 | 27.23 | 26.78 | **26.70** |

According to Table 9, the neural network algorithm had the best performance in most evaluation metrics. In the case of the mean bias error (MBE), the M3 model had the value closest to zero (-0.051) of all models used in the research, followed by the M1 model (0.171). Positive MBE values indicated that the average of the results estimated by the models was greater than the average of the actual observations.

Moreover, the M3 model had a higher performance than the M1 model. In regard to training, the model was better by 5%, and in tests, it was also better by 3% according to the determination coefficient $R^2$, which indicated that the model was better at explaining the variability of the data around the mean and that it fit the data better; in addition, the model had 8% less amount of errors between the real and estimated datasets, according to the RMSE. The M2 model had the lowest performance of the 3 models, although it did not require surface-observed data.

Considering the relative mean square error (rRMSE), the neural networks also had the highest performance; however, the estimated results in the M2 model were higher than 30%, which indicated a poor performance, and in the case of the M1 and M3 models, the forecast results were between 20% and 30%, which represented a regular performance of the machine learning algorithm.

Figure 3 shows the scatter diagrams obtained with the multilayer perceptron technique for each model, using the actual solar radiation captured by the station and the estimated solar radiation by the models for 2012. At first glance, the differences between the three images are not very evident; however, the M3 model fits the data better and represents less variability around the mean.

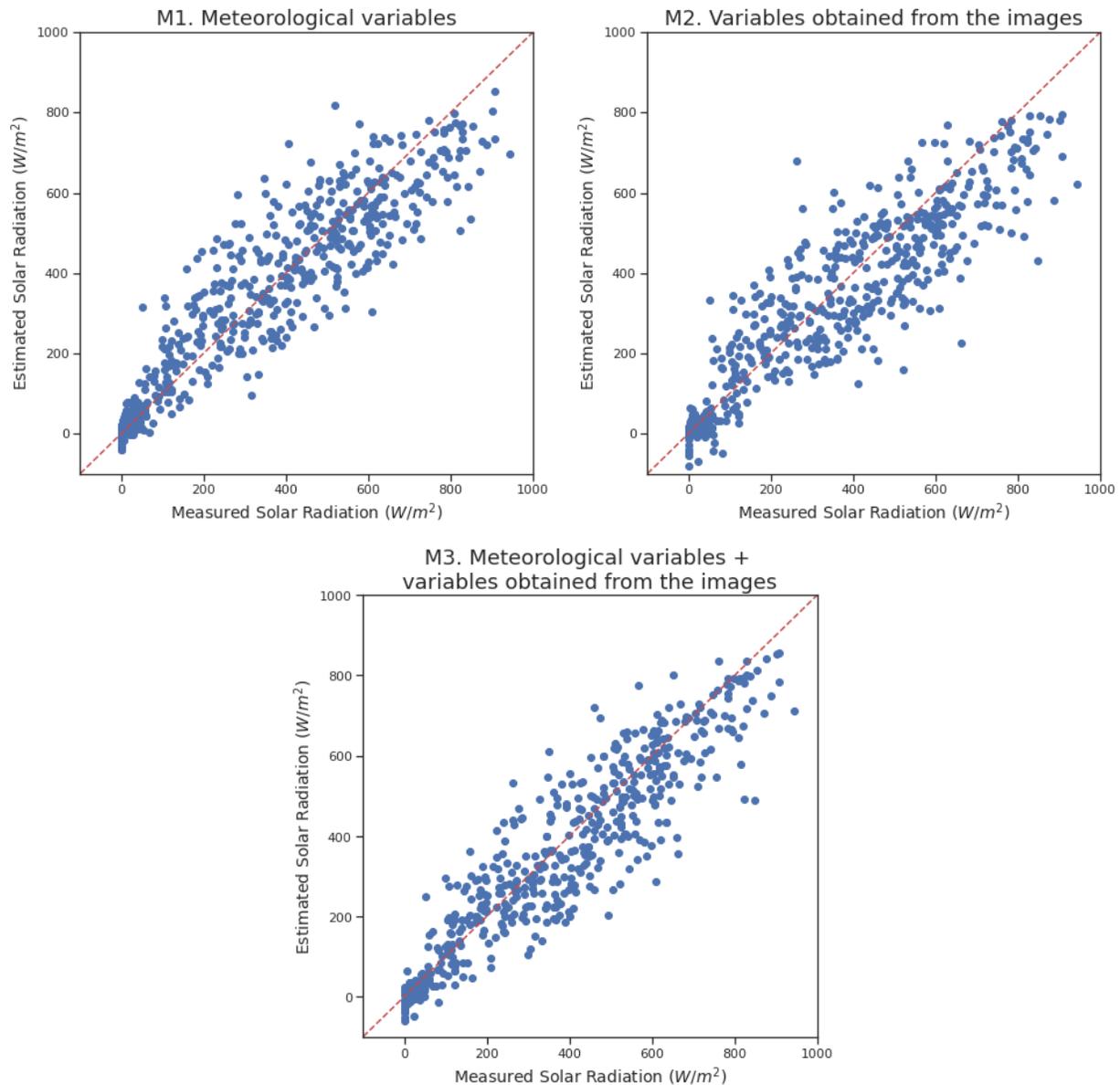

**Figure 3**. Real and estimated solar radiation in 2012 by using the multilayer perceptron.

According to the results in Table 10, the neural network algorithm had the best performance in most evaluation metrics. In the case of the mean bias error (MBE), the M3 model had the closest value to zero (-0.146) of all models used in the research, followed by the M1 model (-0.149). In this case, the MBE value was negative in both cases, which indicated that the average of the results estimated by the models was less than the average of the actual observations.

Moreover, the M3 model also had a higher performance than the M1 model; in training, the model was better by 4%; and in tests, it was also better by 3% according to the determination coefficient

$R^2$, which indicated that the model increased the level in the explanation of the variability of the data around the mean and that it fit the data better. In addition, the model had 6% less amount of errors between the real and estimated datasets, according to the RMSE. The M2 model had the lowest performance of the 3 models, although it did not require surface-observed data.

Analyzing the relative mean square error (rRMSE), the neural networks also had the highest performance, and the results estimated by the three models were between 20% and 30%, which represented a regular performance of the algorithm. The M3 model had the best performance in predictions.

**Table 10**. Results obtained from the 2017 dataset.

| ID | MODEL | METRICS | MULTIPLE LINEAR REGRESSION | GRADIENT BOOSTING | XGBOOST REGRESSOR | RANDOM FOREST | NEURAL NETWORKS |
|---|---|---|---|---|---|---|---|
| M1 | METEOROLOGICAL VARIABLES | MBE | -8.345 | -0.452 | 1.506 | **-0.149** | 1.687 |
| | | $R^2$ ENT | 0.581 | **0.901** | 0.900 | 0.893 | 0.882 |
| | | $R^2$ PRU | 0.611 | 0.870 | 0.867 | 0.866 | **0.890** |
| | | RMSE | 88.48 | 49.03 | 48.19 | 47.20 | **46.97** |
| | | rRMSE | 48.17 | 28.11 | 26.48 | 27.01 | **25.57** |
| M2 | VARIABLES OBTAINED FROM THE IMAGES | MBE | -11.715 | 1.856 | -3.739 | 0.992 | **-0.748** |
| | | $R^2$ ENT | 0.349 | 0.900 | **0.901** | 0.889 | 0.895 |
| | | $R^2$ PRU | 0.325 | 0.872 | 0.868 | 0.863 | **0.881** |
| | | RMSE | 116.58 | 46.73 | **46.28** | 46.36 | 49.04 |
| | | rRMSE | 63.46 | 26.67 | **25.44** | 26.22 | 26.70 |
| M3 | METEOROLOGICAL VARIABLES + VARIABLES OBTAINED FROM THE IMAGES | MBE | -8.994 | -1.076 | -0.298 | 0.580 | **-0.146** |
| | | $R^2$ ENT | 0.657 | **0.933** | 0.925 | 0.916 | 0.927 |
| | | $R^2$ PRU | 0.691 | 0.902 | 0.891 | 0.887 | **0.917** |
| | | RMSE | 78.85 | 44.10 | 47.95 | 42.59 | **40.97** |
| | | rRMSE | 42.92 | 24.73 | 25.74 | 24.92 | **22.30** |

Figure 4 represents the scatter diagrams obtained with the multilayer perceptron technique for each model, using the real solar radiation captured by the station and the estimated solar radiation captured by the models for 2017. Although visually, the differences between the images that represent each model are not very evident, the M3 model fits the data better and represents less variability around the average.

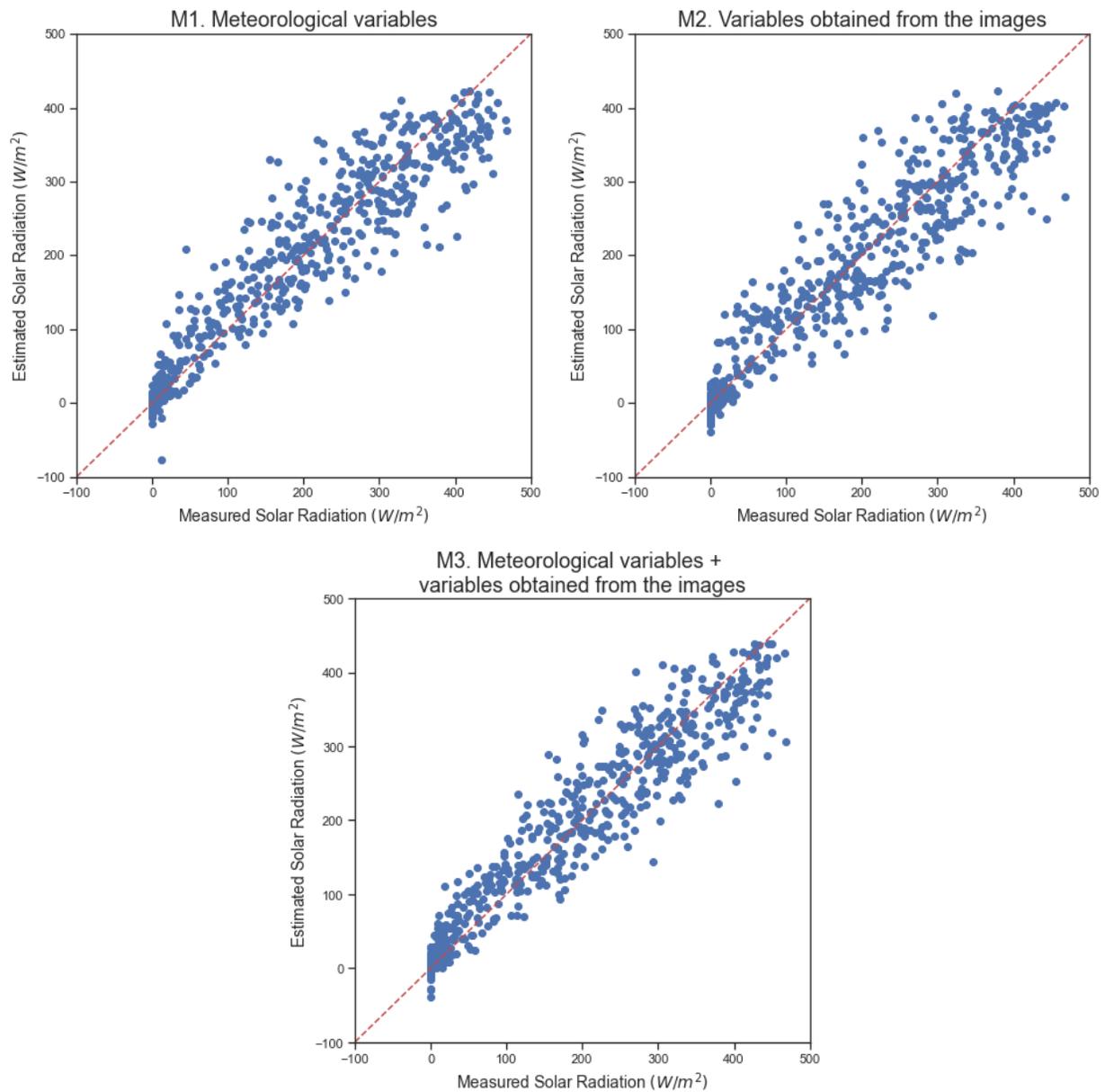

**Figure 4**. Real and estimated solar radiation in 2017 using the multilayer perceptron.

## 4. Discussion

Research on solar radiation predictions occurs in many cases due to the limited number of measurement stations; additionally, because some of them do not provide instruments to measure it, some models that estimate solar radiation from satellite images require radiation and solar brightness data observed on the ground, given that coefficients $a$ and $b$, according to Equation (1), are essential in calculating solar radiation and depend on different geographic and climatic parameters and some atmospheric properties, as observed in the work by Poveda Matallana [11].

This research contributes to the solution of the global problem of energy production associated with the emission of greenhouse gases generated by the burning of fossil fuels, according to the study by Bakay and Ağbulut [47]; Its importance lies in the support given to people who make decisions in the installation of solar farms in remote places because it provides data on the behavior of solar radiation using artificial intelligence techniques based on characteristics obtained from satellite images. This work is unique and valuable because it uses the Angström-Prescott statistical method to extract variables from satellite images integrating them with meteorological data so that they enable the prediction of solar radiation at a given location using machine learning algorithms.

The data acquired from satellite images allow the calculation of variables that directly affect the solar radiation that reaches the Earth's surface; among others, the cloudiness index is a parameter that depends on the minimum and maximum reflectance of each hour of the day. Reflectance values can be above 1, which is why, according to the research by Laguarda et al. [42], only 80% of the maximum reflectance should be used without compromising the performance of the model. Even so, the values of this index must be between 0 and 1; therefore, they must be adjusted in the case of exceeding the limits of the domain.

The low performance of the 2012 statistical model, in part, was due to the lack of images provided by the satellite on certain days of the year and at certain hours of the day; 41.8% of the total images were obtained (1991 out of 4758 possible). In addition, between 6 am and 6 pm, only 7.9% of the days of the year had 13 images; in contrast, 66.4% of the days of the year had 4 images or less, and finally, the parameters observed by the DAGMA air quality stations (ERA and Compartir) did not include the solar brightness variable. Therefore, data from a nearby IDEAM station, the Univalle station, were used, which provided only 89.9% of the daily solar brightness data.

The multiple linear regression model, although it had a lower performance than the other machine learning methods used, was important to this study because it provided an initial idea about the predictions of solar radiation from the selected datasets and the preliminary treatment that was applied to them. Although the neural networks had the highest performance, the assembly methods also showed good performance in relation to their results.

**5. Conclusions**

This paper evaluates the performance of five automatic learning algorithms (multiple linear regression, gradient boosting, XGBoost regressor, random forest and neural networks) to estimate solar radiation in Colombia. The research considers the characteristics obtained from the images taken by the GOES-13 satellite in 2012 and 2017, two datasets provided by air quality measurement stations (Escuela República de Argentina -ERA- and Compartir) from the Mayoralty of Cali and a dataset of daily solar brightness from the Univalle station, supplied by IDEAM.

The time necessary for the request, download, georeferencing and processing of the images to obtain the reflectance of the pixels, given the geographic coordinates, may take approximately 18 hours, depending on the internet connection speed, the availability of the NOAA Class server and the characteristics of the computer equipment on which the statistical model and machine learning algorithms are run.

The reflectance obtained with negative values is produced thanks to the digital level of the images, which is sometimes less than the value of the response instrument in the GOES imager visible channel, to the space scene where the signal is expected to be zero, according to the parameters prior to satellite launch [41], which established this value at 29, as observed in Equation (2).

In the statistical model for 2017, it was not possible to estimate the solar radiation from the images of the GOES-13 satellite due to the absence of solar brightness data at the IDEAM Univalle station, which prevented the calculation of the Angström-Prescott coefficients that, together with the cloudiness index and the theoretical value of solar radiation at the edge of the atmosphere, allow the calculation of solar radiation using Equation (1).

Considering the correlation values in the regression models, the temperature variable, followed by the reflectance variable, had a greater correspondence with solar radiation; similarly, the artificial neural network called the multilayer perceptron, had the best performance in solar radiation predictions compared with the other regression algorithms based on the determination coefficient $R^2$ and the mean square error, RMSE. Furthermore, the model that integrated the meteorological variables and the variables obtained from the satellite images produced the best results in solar radiation estimations in comparison with the statistical model and the other learning models.

The M2 model had a lower performance than the M1 and M3 prediction models; although it provided better results than the statistical model, it also allows the prediction of solar radiation at any geographic location on the planet using only data obtained from the processing of the images recovered from the GOES-13 satellite.

Considering the results obtained with the machine learning algorithms used in this work, we propose acquiring satellite images and recent solar radiation data from at least three consecutive years, with the aim of using time series to make predictions of solar radiation in the future and, if possible, gather as much data as possible so that we can take advantage of the performance of deep learning algorithms.

**Funding**

This research did not receive any specific grant from funding agencies in the public, commercial, or not-for-profit sectors.